\documentclass[runningheads]{llncs}

\usepackage{eccv}

\usepackage{eccvabbrv}

\usepackage{graphicx}
\usepackage{booktabs}
\usepackage{colortbl}
\usepackage{xcolor}
\usepackage{amsmath}
\usepackage{tcolorbox}
\usepackage{multirow}
\usepackage{tikz}

\usepackage[accsupp]{axessibility}  

\usepackage{hyperref}

\usepackage{orcidlink}

\newtcolorbox{redprompt}[1][]{
  colback=red!5!white,     
  colframe=red!80!black,   
  fonttitle=\bfseries,      
  fonttitle=\small,
  fontupper=\scriptsize,
  fontlower=\scriptsize,
  title=AI Prompt,          
  arc=3mm,                  
  boxrule=1pt,              
  #1                        
}

\newtcolorbox{blueprompt}[1][]{
  colback=blue!5!white,     
  colframe=blue!80!black,   
  fonttitle=\bfseries,      
  fonttitle=\small,
  fontupper=\scriptsize,
  fontlower=\scriptsize,
  title=AI Prompt,          
  arc=3mm,                  
  boxrule=1pt,              
  #1                        
}

\begin{document}

\title{BEiTScore: Reference-free Image Captioning Evaluation with an Efficient Cross-Encoder Model} 


\author{Gonçalo Gomes\inst{1,2,3}\orcidlink{0009-0007-3199-8992} \and
Bruno Martins\inst{1,2}\orcidlink{0000-0002-3856-2936} \and
Chrysoula Zerva\inst{3}\orcidlink{0000-0002-4031-9492}}

\authorrunning{G.~Gomes et al.}

\institute{Instituto Superior Técnico, University of Lisbon \and
INESC-ID \and Instituto de Telecomunicações \\
\email{\{goncaloecgomes, chrysoula.zerva, bruno.g.martins\}@tecnico.ulisboa.pt}}

\maketitle

\begin{abstract}
Image captioning evaluation remains a significant challenge, as vision-language models evolve toward more challenging capabilities such as generating long-form and context-rich descriptions. State-of-the-art evaluation metrics involve extensive computational costs associated with the use of Large Language Models (LLMs) as judges, or instead suffer from the limitations of standard CLIP-based encoders, such as strict token limits, lack of fine-grained sensitivity, or lack of compositional generalization by treating captions as ``bags-of-words.'' We propose a new learned metric that tackles the aforementioned challenges, based on a lightweight cross-encoder that is initialized from a visual question-answering model checkpoint, balancing a strong weight initialization with computational efficiency. Our training scheme uses a carefully assembled data mixture for supervised learning, featuring adversarial LLM-based data augmentations to enhance model sensitivity to fine-grained visual-linguistic errors. We also introduce a new benchmark designed to assess detailed captioning evaluation across diverse scenarios. Experimental results demonstrate that the proposed metric achieves state-of-the-art performance while maintaining the efficiency required for large-scale benchmarking, quality-aware decoding, or reward guidance.

\end{abstract}

\section{Introduction}
\label{sec:intro}

\begin{figure}
    \centering
    \vspace{-1mm}
    \includegraphics[width=0.70\linewidth]{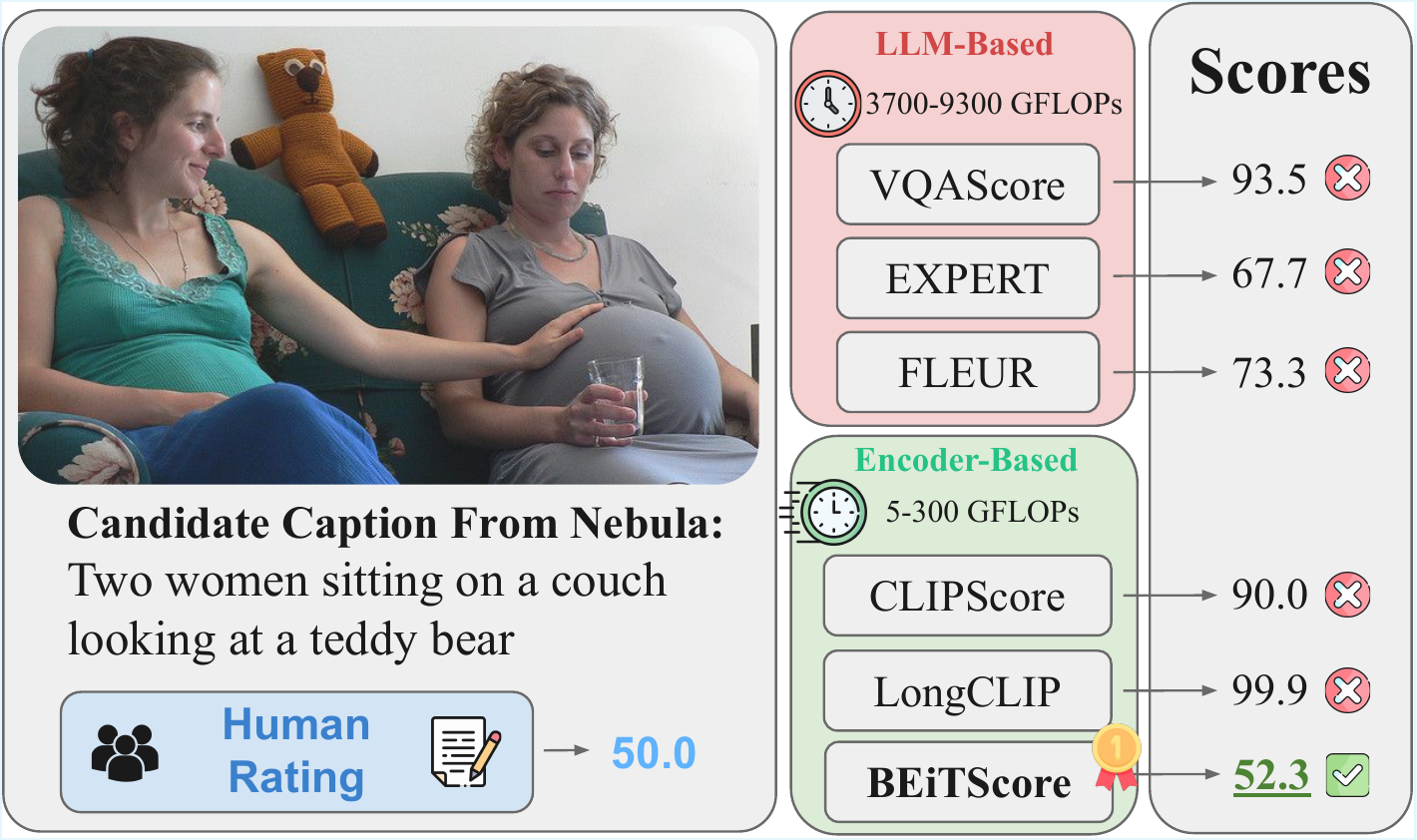}
    \caption{BEiTScore versus state-of-the-art metrics on an instance from Nebula~\cite{matsuda2024deneb}.}
    \vspace{-5mm}
    \label{fig:placeholder}
\end{figure}
The combination of the visual and linguistic modalities has developed as a significant research frontier, with image captioning serving as a foundational task for Vision and Language Models (VLMs). In this context, the development of robust evaluation metrics, considering cost-effective architectures, is critical, both for benchmarking VLM performance and for guiding caption generation, whether by quality-aware decoding or through reward-based guidance.

For image captioning evaluation metrics to be useful, they must achieve high correlation with human judgments while handling long-form descriptions and detecting fine-grained visual-linguistic errors. In addition, they should ideally be able to evaluate captions without relying on human-written references. While there have been significant advances in terms of learned evaluation metrics, the current encoder-based metrics still struggle with these requirements. Recent encoder-based approaches either rely on reference captions to assist evaluation within the textual domain~\cite{wada2024polos,matsuda2024deneb,chen2025specs} or use CLIP~\cite{hessel2021clipscore}-based encoders. Despite achieving strong correlation with human judgments~\cite{hessel2021clipscore}, CLIP-based metrics are also limited to processing modalities independently, and they often see captions as bags-of-words~\cite{yuksekgonul2023and}, thus failing to capture subtle relationships between vision-linguistic content, as well as over lengthy descriptive narratives~\cite{urbanek2024picture}. Furthermore, due to limited exposure to long-form captions during pretraining, CLIP models are typically biased toward short captions and struggle to process descriptions beyond the original 77-token limit, leading to poor performance on context-rich content~\cite{urbanek2024picture, feng2025retaining}.

Recently, LLMs have attracted attention for their zero-shot and long context capabilities in image captioning evaluation~\cite{chan2023clair, lee2024fleur, kim2025expert}, however, they also bring deployment challenges. For instance, their architectural complexity leads to long inference times, limiting their usefulness for model development and large-scale benchmarking. Such processing costs cannot be ignored, and preserving a balance between evaluation quality and computational efficiency is important.

To address these limitations, we propose a new learned metric that leverages a lightweight cross-attention encoder model, using the BEiT-3~\cite{wang2023image} architecture, initialized from a Vision Question Answering (VQA) checkpoint. Our training scheme uses a data mixture specifically designed to increase the model’s exposure to fine-grained descriptions that capture diverse visual–linguistic relationships. Our synthetic data generation pipeline incorporates specialized, adversarial LLM-driven augmentations to systematically enrich compositional coverage and relational complexity. In addition, we present a new benchmark for evaluating long-form, detail-oriented captioning with an emphasis on vision–language compositional reasoning. The benchmark spans a broad spectrum of visual–textual relations, including in-image text recognition, an aspect largely underrepresented in prior work. Finally, we validate our approach across established benchmarks, demonstrating consistent improvements against state-of-the-art models. 

Results indicate that prior encoder-based models struggle to process long captions effectively. In contrast, BEiTScore preserves robust compositional understanding across both short and long descriptions, while excelling in alignment with human judgments. It matches or surpasses the performance of specialized learned LLM-based metrics across all of the considered benchmarks in Section~\ref{sec:results}, while remaining $20\times \text{ to } 60\times$ lighter and $30\times \text{ to }100\times$ faster at inference time.

\section{Related Work}

This section reviews relevant prior work, organized into three key areas: BEiT-3, image captioning evaluation, and benchmarking of VLM evaluation metrics.

\subsection{The BEiT-3 Encoder Model}

BEiT-3~\cite{wang2023image} was introduced as a unified multimodal foundation model that represents images through discrete tokens analogous to language tokens, allowing for a single learning framework that masks and predicts both visual and textual inputs. By treating image-text pairs as instances of a shared input, BEiT-3 learns cross-modal representations within a unified token space. Its architecture corresponds to a multiway Transformer that shares self-attention parameters across modalities, with feed-forward layers that specialize to either text or vision processing. The resulting framework demonstrates good generalization capabilities and achieves strong performance across a broad spectrum of vision-language tasks, including image classification, visual reasoning, and image captioning.

\subsection{Image Captioning Evaluation}

CLIPScore~\cite{hessel2021clipscore} pioneered the approach of performing image captioning evaluation without the use of human-written captions as references. This method consists of computing a scaled cosine similarity between image and caption embeddings produced by a CLIP dual encoder to assess alignment, showing strong performance on short visually verifiable content, and inspiring several subsequent learned metrics with CLIP-based architectures~\cite{sarto2025positive,hu2023infometic,narins2023validated,wada2024polos,gomes2025evaluation}, including versions that extend CLIP to support long-text input~\cite{chen2025specs, zhang2024long}.
Other approaches instead relied on the joint encoding of images and texts. For instance, VQAScore~\cite{lin2024evaluating} uses a large VQA model to produce an alignment score by computing the probability of a ``\texttt{Yes}'' answer to a simple ``\texttt{Does this figure show $<text>$?}'' question.  

Recent advances have shown that LLM-based metrics often surpass CLIP-based approaches in human alignment. Early efforts such as CLAIR~\cite{chan2023clair} used LLMs as reference-based judges, but this method operated purely in the textual domain and lacked guarantees regarding output-format consistency. FLEUR~\cite{lee2024fleur} transitioned to reference-free evaluation by conditioning directly on the image. EXPERT~\cite{kim2025expert} further refined previous approaches by fine-tuning VLMs specifically for scoring, improving reliability and output consistency. However, the architectural complexity of LLM-based metrics leads to long inference times, limiting their usefulness for model development and large-scale benchmarking.

\subsection{Benchmarking VLMs and VLM Evaluation Metrics}

For image captioning evaluation metrics, the most critical property is correlation with human judgments. An effective metric should reflect how well a generated caption aligns with human perceptions of image-text correspondence. To support this goal, previous work has introduced several datasets and benchmarks that provide explicit human ratings of image-caption alignment quality. These datasets enable quantitative comparison between automatic metrics and human ratings with some containing both training and benchmarking versions. Notable examples include Flickr8K-EX/CF~\cite{hodosh2013framing}, Composite~\cite{aditya2015images},  VICR~\cite{narins2023validated}, POLOS~\cite{wada2024polos}, Nebula~\cite{matsuda2024deneb}, and EvalMuse-40k~\cite{han2024evalmuse}, which differ in annotation protocols, rating scales, and emphasis on robustness to hallucinations or semantic similarity. 

Although extensively used, the previous datasets predominantly emphasize overall alignment quality and are typically limited to short captions. Consequently, they lack coverage of complex semantic phenomena and fine-grained perception capabilities. In particular, they do not systematically assess a metric’s sensitivity to fine-grained image-text relationships nor its ability to properly capture spatial relations, compositional structure, negations, attribute binding, word order sensitivity, counting, action understanding, hallucination detection, or role consistency (e.g., agent-patient swaps). Yet, the ability to capture and discriminate such relationships is critical for image captioning metrics, and existing correlation-based benchmarks largely overlook these factors, which makes the performance of metrics on complex image-text interactions difficult to interpret.

Therefore, to complement correlation-based benchmarks, metrics are often evaluated using targeted binary or contrastive benchmarks designed to probe specific linguistic and visual perception skills~\cite{hsieh2023sugarcrepe}. Benchmarks such as FOIL-It~\cite{shekhar2017foil}, nocaps-FOIL~\cite{petryk2024aloha}, VALSE~\cite{parcalabescu2022valse}, SUGARCREPE~\cite{hsieh2023sugarcrepe}, or Winoground~\cite{thrush2022winoground}, all make use of minimally perturbed caption pairs that differ along controlled semantic dimensions, enabling precise measurement of sensitivity to changes in relational understanding, attribute grounding, and compositional semantics. While these contrastive benchmarks do not directly measure human judgment correlation, they provide critical complementary signals by revealing whether an evaluation metric captures deeper aspects of visual-linguistic understanding beyond surface-level similarity. However, despite these benchmarks probing diverse linguistic and visual reasoning skills, they still focus on short captions.

\section{Training BEiT-3 For Image Captioning Evaluation}

The most widely used image captioning datasets (e.g., MS-COCO~\cite{lin2014microsoft}) predominantly feature short captions, emphasizing concise object-level descriptions rather than rich and detailed semantic coverage. Consequently, smaller vision-language models trained on such data are rarely exposed to highly descriptive texts. We argue that small encoder-based models can effectively support the evaluation of detailed captions upon sufficient exposure to more complex and long-form textual data during training. 
\begin{figure}[t]
    \centering
    \includegraphics[width=1\linewidth]{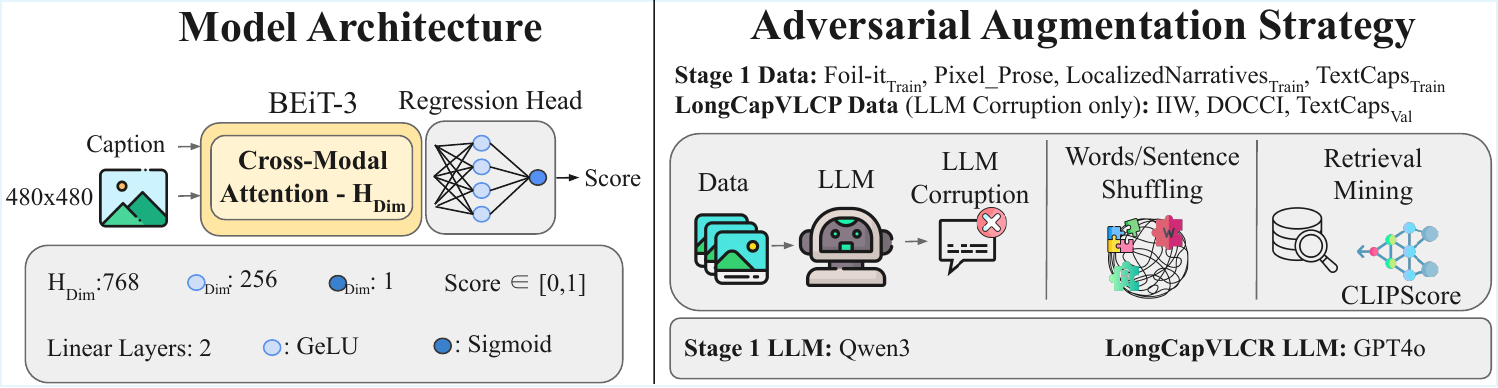}
    \caption{BEiTScore architecture, and adversarial data augmentation strategy.}
    \label{fig:placeholder}
\end{figure}

\subsection{Adversarial Data Augmentation Strategy}

Building on the aforementioned idea, we augmented existing datasets by explicitly focusing on pairing images with detailed textual descriptions~\cite{shekhar2017foil,pont2020connecting, singla2024pixels, sidorov2020textcaps}, and repurposing them to build a binary pairwise training dataset. Specifically, we prompt a LLM\footnote{We used Qwen3 (https://huggingface.co/Qwen/Qwen3-30B-A3B-Instruct-2507) to augment our training data, and GPT4o to generate our benchmark.} to generate factually incorrect but fluent captions based on the original (correct) captions, by modifying specific textual elements (i.e., objects and attributes) and diverse visual-linguistic relations (i.e., spatial relations, actions, role consistency, or counting). This strategy creates detailed contrastive examples that explicitly train image-text alignment and perception.

In addition to LLM-based caption perturbations, we employ a synthetic augmentation strategy based on text corruption, inspired by the work of Ma et al.~\cite{ma2024cobra}. Prior work has shown that many CLIP-based metrics exhibit bag-of-words behavior~\cite{yuksekgonul2023and, ma2024cobra}, where repeating keywords or reordering phrases can artificially inflate similarity scores without improving true semantic alignment. To counteract this, we synthetically generate incorrect captions by corrupting the original ground-truth text, targeting repetitions of specific parts of speech (e.g., nouns, verbs, adjectives, adverbs) and random word and sentence shuffling. 

Finally, we incorporate a retrieval-based hard negative mining strategy. For each image-caption pair in a set of collected datasets, we used a retrieval model (beit3\_large\_itc\footnote{https://github.com/microsoft/unilm/blob/master/beit3/README.md}) and select incorrect captions that are nevertheless assigned high similarity. This enables us to stratify negatives into \texttt{easy}, \texttt{medium}, and \texttt{hard} categories, using three different ranges of similarity scores found qualitatively. 

\subsection{Training Scheme}

Training uses two distinct phases with two different categories of data:
\begin{itemize}\setlength{\itemsep}{0pt}
    \item \textbf{Pairwise Contrastive Data:} This set is constructed using our custom augmentation strategy applied to Foil-It~\cite{shekhar2017foil}, a short-caption pairwise dataset, TextCaps~\cite{sidorov2020textcaps}, which is a dataset focusing on scene text analysis, as well as to long-description datasets. Specifically, for long descriptions, we use PixelProse~\cite{singla2024pixels}, and LocalizedNarratives~\cite{pont2020connecting}. This augmented mixture of datasets require the model to perform binary discrimination, identifying which of two provided captions is factually consistent with a given image, thus supporting pairwise contrastive training.
    \item \textbf{Human Correlation Data:} This includes the training splits of datasets such as VICR~\cite{narins2023validated}, POLARIS~\cite{wada2024polos}, NEBULA~\cite{matsuda2024deneb}, and EvalMuse-40k~\cite{han2024evalmuse}, where image-caption pairs were annotated with continuous alignment scores derived from human judgments $q_i$, which we normalize so that $q_i \in [0,1], \forall i$.
\end{itemize}
Note that both the training and validation sets were strictly derived from the respective official training and validation splits of each constituent dataset.
The pairwise contrastive data serves as a discretized proxy for human rating regression. Consequently, training with these data provides a robust initialization for subsequent fine-tuning. In addition, the inclusion of pairwise contrastive data exposes the training model to longer, more descriptive captions and richer image-text relationships that complement the specialized human correlation datasets.
Consequently, our proposed training process consists of two sequential stages.
\begin{enumerate}
    \item[S1] \textbf{Pairwise Contrastive Learning:} The model is trained to distinguish correct from incorrect captions using a Binary Cross-Entropy (BCE) loss. To optimize feature acquisition, we adopt an iterative training schedule:
    \begin{enumerate}
        \item \textbf{Foundational Reasoning:} Initial training focuses on a core set of linguistic and visual errors, including spatial relations, role consistency, action verbs, nouns, and adjectives.
        \item \textbf{Quantitative Refinement:} Building upon the checkpoint from S1.a, we introduce counting-related errors to the training mixture.
        \item \textbf{Text in Image Integration:} In the final iteration of S1, we incorporate into the training mixture errors regarding scene-text information appearing within the image.
    \end{enumerate}
    \item[S2] \textbf{Alignment Fine-tuning:} The model is fine-tuned directly on human judgment datasets. We employ an $L_1$ loss to optimize for direct alignment with continuous human scores, transitioning the model from binary classification to fine-grained quality estimation.
\end{enumerate}

We empirically observed that the incremental expansion of the training distribution in S1 yields superior performance compared to training on the entire data mixture simultaneously. Inspired by the VQAScore approach~\cite{lin2024evaluating}, we warm-started the model from the BEiT-3 VQA checkpoint, as this pretraining provides useful prior knowledge for our objective, while leveraging cross-modal attention to jointly process vision and language information effectively. We trained two BEiT-3 model variants, BEiT-3 ViT-B/32 with $222$ billion parameters, and BEiT-3 ViT-L/14 with $674$ billion parameters. We applied a drop path rate of 15\% during training for regularization. The regression head consists of a single hidden layer with $256$ units, GeLU activations, and a single-node output layer with a sigmoid activation to produce the final score. All models were trained on a single NVIDIA H200 GPU using a batch size of 24 and a learning rate of $5×10^{-7}$.\looseness-1

\section{Experimental Results}
\label{sec:results}

This section details the benchmark datasets, quantitative analyses, and comparative evaluations made with the proposed metric against other state-of-the-art methods. Our evaluation focuses exclusively on reference-free image captioning metrics, as we believe that reliance on reference captions is impractical for most real-world applications, where human-annotated ground truth is rarely available. Performance is assessed via correlation with human judgments and fine-grained visual-linguistic perception, complemented by qualitative analyses.

\subsection{Datasets}

To ensure a fair and comprehensive evaluation of the correlation between the proposed metric and human judgments, we conduct experiments on several different benchmark datasets featuring image-caption pairs associated to numeric ratings. These include VICR~\cite{narins2023validated}, Polos~\cite{wada2024polos}, Nebula~\cite{matsuda2024deneb}, Flickr8KExpert~\cite{hodosh2013framing}, Flickr8K-CF~\cite{hodosh2013framing}, and Composite~\cite{aditya2015images}. Furthermore, to assess the robustness of our model across specific linguistic and visual perception capabilities, we evaluate performance on several targeted diagnostic benchmarks, including nocaps-FOIL~\cite{petryk2024aloha}, SugarCREPE~\cite{hsieh2023sugarcrepe}, VALSE~\cite{parcalabescu2022valse}, and Winoground~\cite{thrush2022winoground}. 

While the aforementioned datasets are gold standards for evaluating fine-grained visio-linguistic capabilities, they possess a significant structural limitation: they are designed for short-form captions. Most feature captions within a 77-token limit, or place critical discriminative errors early in the text. To bridge this gap, we developed a new benchmark named LongCapVLCP, designed to assess long-form narrative captioning beyond the 77-token threshold. In addition, we broaden the evaluation scope by introducing a new task that probes whether existing state-of-the-art image captioning metrics can effectively support scene text perception, a capability largely overlooked by prior VL benchmarks.

Our benchmark leverages high-quality data from three foundational sources: Image In Words~\cite{garg2024imageinwords} and DOCCI~\cite{onoe2024docci} for detailed descriptive grounding, and the validation split of TextCaps~\cite{sidorov2020textcaps} to assess scene text understanding.

To generate challenging incorrect captions, while avoiding the use of the same LLM that was used for constructing our training set, we used GPT-4o with the image and original caption as inputs, prompting it to synthesize negative examples for each image-caption pair, preserving overall fluency. These ``hard negatives'' fall into the following error categories:
\begin{itemize}
    \item \textbf{Actant Swap:} Swapping agent-patient roles in the caption. 
    \item \textbf{Action Verbs:} Altering the primary actions or dynamics within the scene.
    \item \textbf{Counting:} Modifying the quantity of specific entities. 
    \item \textbf{Nouns/Adjectives:} Altering objects and attributes present in the caption. 
    \item \textbf{Spatial Relations:} Modifying the spatial relations present in the caption. 
    \item \textbf{Scene Text:} Add errors related to the textual information within the image. 
\end{itemize}

\subsection{Results on Correlation with Human Judgments}

\begin{table}[t]
\centering
\small
\resizebox{\columnwidth}{!}{%
\begin{tabular}{l c c c c c c c c c}
\toprule
\textbf{Model} &\textbf{Size (M)} & \textbf{GFLOPs} & & \textbf{VICR$\tau_c$} & \textbf{Polaris$\tau_c$} & \textbf{Nebula$\tau_c$} & \textbf{EX-8K$\tau_c$} & \textbf{CF-8K$\tau_b$} & \textbf{COM$\tau_c$} \\
\midrule
\rowcolor{gray!15}
\multicolumn{10}{c}{\textit{LLM-based Methods}} \\
\midrule
FLEUR~\cite{lee2024fleur}   & $13350.84$ & $9317.47$ & & $74.7$ & $58.4$ & $52.7$ & $53.0$ & $38.6$ & \underline{60.8} \\
EXPERT~\cite{kim2025expert}    & $13350.84$ & $9317.47$ & & $77.4$ & $60.9$ & $54.9$ & $56.7$ & \underline{39.3} & \textbf{61.8} \\
VQAScore (CLIP-FlanT5-XXL)~\cite{lin2024evaluating} & $11459.59$ & $3742.1$ & & $69.8$ & $51.6$ & $49.4$ & $41.0$ & $36.5$ & $53.4$ \\
\midrule
\rowcolor{gray!15}
\multicolumn{10}{c}{\textit{Encoder-based Methods}} \\
\midrule
CLIP-S/ViT-B~\cite{hessel2021clipscore} & $151.28$ & $7.28$ & & $70.8$ & $52.9$ & $47.8$ & $52.3$ & $35.3$ & $51.8$ \\
CLIP-S/ViT-L~\cite{hessel2021clipscore} & $427.62$ & $84.36$ & & $72.6$ & $53.2$ & $48.8$ & $54.7$ & $36.5$ & $55.4$ \\
LongCLIP-S/ViT-B~\cite{zhang2024long}   & $149.84$ & $26.24$ & & $70.2$ & $54.8$ & $48.9$ & $53.0$ & $35.2$ & $51.3$ \\
LongCLIP-S/ViT-L~\cite{zhang2024long}   & $427.75$ & $98.9$ & & $69.0$ & $53.9$ & $48.6$ & $53.9$ & $35.9$ & $50.5$ \\
PAC-S/ViT-B~\cite{sarto2023positive}    & $151.28$ & $4.88$ & & $71.4$ & $52.3$ & $47.2$ & $54.3$ & $36.0$ & $53.5$ \\
PAC-S/ViT-L~\cite{sarto2023positive}    & $427.62$ & $56.26$ & & $73.2$ & $53.1$ & $47.9$ & $55.7$ & $37.0$ & $55.4$ \\
PAC-S++/ViT-B~\cite{sarto2025positive}  & $155.42$ & $4.88$ & & $72.7$ & $52.4$ & $46.9$ & $54.5$ & $37.0$ & $55.7$ \\
PAC-S++/ViT-L~\cite{sarto2025positive}  & $430.58$ & $56.26$ & & $75.2$ & $53.6$ & $48.9$ & $57.4$ & $38.5$ & $59.4$ \\
SPECS~\cite{chen2025specs}              & $151.49$     & $9.2$  &  & $33.4$ & $39.7$ & $34.4$ & $30.4$ & $18.2$ & $24.1$ \\
\midrule
BEiTScore-B/32 w/o Stage 1 & $222.28$ & $83.85$ & & $77.7$ & $60.9$ & $54.6$ & $57.6$ & $37.8$  & $59.2$  \\
BEiTScore-L/14 w/o Stage 1 & $674.69$ & $296.73$ & & \underline{78.3} & \underline{62.2} & \underline{56.3} & $57.8$ & $38.5$  & $59.5$  \\
BEiTScore-B/32                & $222.28$ & $83.85$ & & $78.2$ & $61.5$ & $55.3$ & \underline{59.4} & $38.5$  & $58.9$  \\
BEiTScore-L/14                & $674.69$ & $296.73$ & & \textbf{78.8} & \textbf{63.2} & \textbf{56.7} & \textbf{60.2} & \textbf{39.6}  & $59.5$  \\
\bottomrule
\end{tabular}}
\caption{Comparison against state-of-the-art reference-free metrics across multiple human correlation benchmarks. Methods are categorized as encoder-based or LLM-based. Performance is measured using Kendall’s $\tau_c$ correlation with human judgments on all benchmarks; for CF-8K, we follow prior work and report Kendall’s $\tau_b$. The highest score is highlighted using \textbf{bold}, and the second best is highlighted using an \underline{underline}.}
\label{tab:human_corr}
\vspace{-5mm}
\end{table}
We first evaluated the proposed method alongside existing state-of-the-art models across six benchmarks, measuring Kendall-tau correlation with human judgments. The results can be seen in Table~\ref{tab:human_corr}. Compared to encoder-based approaches, both our BEiTScore-B/32 and BEiTScore-L/14 variants surpass previous state-of-the-art methods across all benchmarks by a significant margin, exceeding 9 points in correlation in some cases, while maintaining comparable model sizes and computational costs in terms of GFLOPs. 

Comparing results against LLM-based methods, our models outperform VQAScore, FLEUR, and EXPERT across nearly all benchmarks. Even for EXPERT, an LLM-based metric fine-tuned for image captioning evaluation, we observe gains in some cases exceeding 7 points in correlation. The only exception is the COM benchmark, where our models still achieve competitive performance. These results are achieved despite our models being smaller (20 to 60 times), and more computationally efficient (10 to 100 times) than their LLM-based counterparts.

It is noteworthy that when considering performance solely on the human correlation benchmarks, the intermediate Stage 1 (S1) of training appears redundant, as the results with and without binary pairwise training are nearly identical. This observation aligns with our initial hypothesis that existing human correlation benchmarks predominantly emphasize global alignment with relatively short captions, and we assert that relying exclusively on these benchmarks is insufficient for a comprehensive assessment of image captioning metrics.

\subsection{Results on Object Hallucination}

\begin{table}[t]
\centering
\small
\resizebox{0.9\columnwidth}{!}{%
\begin{tabular}{l c c c c c c c c c c c c c c c}
\toprule
  & \multicolumn{3}{c}{\textbf{Overall}} && \multicolumn{3}{c}{\textbf{In-Domain}} && \multicolumn{3}{c}{\textbf{Near-Domain}} && \multicolumn{3}{c}{\textbf{Out-Domain}} \\
\cmidrule(lr){2-4} \cmidrule(lr){6-8} \cmidrule(lr){10-12} \cmidrule(lr){14-16}
 & AP & AUC & ACC && AP & AUC & ACC && AP & AUC & ACC && AP & AUC & ACC \\
\midrule
\rowcolor{gray!15}
\multicolumn{16}{c}{\textit{LLM-based Methods}} \\
\midrule
ALOHa$\dagger$~\cite{petryk2024aloha} & $69.3$ & $--$ & $--$ && $71.8$ & $--$ & $--$ && $66.7$ & $--$ & $--$ && $70.9$ &  $--$ & $--$  \\
FLEUR~\cite{lee2024fleur}    & $86.8$ & $88.7$ & $97.2$ && $82.5$ & $86.6$ & $96.3$ && $88.4$ & \underline{90.0} & $97.9$ && $85.2$ & $87.0$ & $96.2$ \\
EXPERT~\cite{kim2025expert}  & \textbf{91.1} & \textbf{92.2} & \textbf{98.5} && \textbf{88.8} & \textbf{91.7} & \underline{97.6} && \textbf{92.6} & \textbf{93.0} & \underline{98.5} && \textbf{89.1} & \textbf{90.8} & \textbf{98.7} \\
GPT4o (API) & $--$ & $--$ & $92.9$ && $--$ & $--$ & $94.7$ && $--$ & $--$ & $93.6$ && $--$ &  $--$ & $90.9$  \\
VQAScore (CLIP-FlanT5-XXL)~\cite{lin2024evaluating} & $88.0$ & $\underline{89.1}$ & $\underline{95.2}$ && $\underline{88.5}$ & $90.2$ & $95.9$ && $\underline{88.8}$ & $89.4$ & $95.1$ && $86.8$ &  $87.7$ & $94.6$  \\
\midrule
\rowcolor{gray!15}
\multicolumn{16}{c}{\textit{Encoder-based Methods}} \\
\midrule
CLIP-S/ViT-B~\cite{hessel2021clipscore} & $70.0$ & $73.7$ & $87.5$  && $67.0$  & $70.0$  & $84.1$ && $68.7$ & $71.8$ & $87.0$ && $73.8$ & $78.6$ & $89.5$ \\
CLIP-S/ViT-L~\cite{hessel2021clipscore} & $71.7$ & $75.1$ & $90.7$  && $69.0$  & $71.8$  & $87.8$ && $70.8$ & $73.9$ & $90.3$ && $74.6$ & $78.2$ & $92.5$ \\
LongCLIP-S/ViT-B~\cite{zhang2024long}   & $74.1$ & $77.1$ & $95.0$  && $72.8$  & $76.1$  & $93.1$ && $72.9$ & $75.9$ & $94.9$ && $77.0$ & $79.6$ & $95.8$ \\
LongCLIP-S/ViT-L~\cite{zhang2024long}   & $78.2$ & $80.0$ & $96.8$  && $74.9$  & $76.8$  & $97.1$ && $76.9$ & $78.8$ & $96.3$ && $81.9$ & $83.4$ & $97.5$ \\
PAC-S/ViT-B~\cite{sarto2023positive}    & $74.7$ & $78.2$ & $91.8$ && $71.3$ & $74.6$ & $89.4$ && $73.2$ & $76.9$ & $90.7$ && $78.7$ & $81.9$ & $94.6$ \\
PAC-S/ViT-L~\cite{sarto2023positive}    & $74.7$ & $78.3$ & $92.7$ && $73.5$ & $77.0$ & $91.4$ && $74.2$ & $77.8$ & $92.5$ && $76.2$ & $79.9$ & $93.4$ \\
PAC-S++/ViT-B~\cite{sarto2025positive}  & $74.2$ & $77.9$ & $92.8$ && $71.9$ & $74.7$ & $90.6$ && $73.2$ & $77.0$ & $92.7$ && $76.9$ & $80.7$ & $93.7$ \\
PAC-S++/ViT-L~\cite{sarto2025positive}  & $77.5$ & $81.0$ & $95.7$ && $73.3$ & $77.8$ & $95.5$ && $77.0$ & $80.7$ & $95.1$ && $79.8$ & $82.7$ & $97.1$ \\
SPECS~\cite{chen2025specs}              & $71.1$ & $74.3$ & $97.0$ && $73.1$ & $76.5$ & $\textbf{98.0}$ && $71.7$ & $75.1$ & $97.3$ && $69.8$ &  $72.2$ & $96.2$  \\
\midrule
BEiTScore-B/32 w/o Stage 1 & $70.4$ & $73.1$ & $94.1$ && $70.0$ & $72.9$ & $94.3$ && $69.8$ & $72.6$ & $94.2$ && $73.5$ & $75.1$ & $94.0$ \\
BEiTScore-L/14 w/o Stage 1 & $72.5$ & $75.5$ & $95.8$ && $71.4$ & $76.3$ & $94.3$ && $71.9$ & $74.4$ & $96.2$ && $76.2$ & $78.5$ & $95.5$ \\
BEiTScore-B/32                & $85.5$ & $86.3$ & $98.1$ && $85.8$ & $87.7$ & \textbf{98.0} && $86.1$ & $86.8$ & \textbf{98.6} && $84.5$ & $85.1$ & $97.2$ \\
BEiTScore-L/14                & \underline{87.3} & $88.8$ & $98.2$ && $87.9$ & \underline{90.7} & $97.1$ && $87.5$ & $88.9$ & $98.3$ && \underline{87.0} & \underline{88.0} & \underline{98.3} \\
\bottomrule
\end{tabular}}
\caption{Comparison against state-of-the-art reference-free metrics across the nocaps-FOIL~\cite{petryk2024aloha} benchmark. Methods are categorized as encoder-based or LLM-based. Performance is measured using Average Precision (AP), Area Under the Curve of the Receiver Operating Characteristic (AUC), and Accuracy (ACC). The highest score is highlighted using \textbf{bold}, and the second best is highlighted using an \underline{underline}. The results for models marked with $\dagger$ were taken from the original papers.}
\label{tab:object_hallu}
\vspace{-5mm}
\end{table}

We evaluated the proposed method against state-of-the-art models on the nocaps-FOIL~\cite{petryk2024aloha} benchmark, which assesses a model's ability to discriminate between correct captions and those containing object hallucinations, where nouns in the ground-truth caption are replaced by semantically related but contextually incorrect objects (e.g., replacing ``car'' with ``bus''). Originally introduced by FOIL~\cite{shekhar2017foil} for MS-COCO, the object replacement methodology was extended by Petryk et al.~\cite{petryk2024aloha} to incorporate captions from the nocaps dataset~\cite{agrawal2019nocaps}, containing objects from classes beyond the MS-COCO~\cite{lin2014microsoft} domain. 

The results in Table~\ref{tab:object_hallu} indicate that both the BEiTScore-B/32 and BEiTScore-L/14 variants of our method surpass existing encoder-based models and also the LLM-based model proposed in the original benchmark. While the accuracy of our method is similar to recent encoder-based approaches, the improvements in AP and AUC are significantly more notable. This suggests that our method achieves superior linear separability between correct and incorrect captions. Furthermore, our method achieves performance comparable to state-of-the-art LLM-based methods, despite being significantly smaller and more efficient.

An ablation of our training strategy highlights the benefits of the Stage 1 pre-training objective (compare to w/o Stage 1 versions in Table \ref{tab:human_corr}). We observe stronger results when the model is gradually fine-tuned on our curated dataset prior to performing human-judgment regression.

\subsection{Results on Object and Attribute Hallucination}

\begin{table}[t]
\centering
\small
\resizebox{0.9\columnwidth}{!}{%
\begin{tabular}{l c c c c c c c c c c c}
\toprule
 & && \multicolumn{2}{c}{\textbf{Add}} && \multicolumn{3}{c}{\textbf{Replace}} && \multicolumn{2}{c}{\textbf{Swap}} \\
  \cmidrule(lr){4-5} \cmidrule(lr){7-9} \cmidrule(lr){11-12}
 & Avg-M. && Attribute & Object && Attribute & Object & Relation && Attribute & Object \\
\midrule
Human$\dagger$   & $98.86$ && $99.0$ & $97.0$ && $99.0$ & $100$ & $99.0$ && $99.0$ & $99.0$ \\
\midrule
\rowcolor{gray!15}
\multicolumn{12}{c}{\textit{LLM-based Methods}} \\
\midrule
FLEUR~\cite{lee2024fleur}    & $88.0$ && $88.0$ & $87.5$ && $93.7$ & $91.0$  & \underline{87.6} && $88.7$ & $86.5$ \\
EXPERT~\cite{kim2025expert}  & \underline{89.7} && \underline{89.2} & $89.7$ && \textbf{95.6} & $89.7$  & $86.5$ && $90.5$ & \textbf{89.8} \\
GPT4o (API)  & $66.7$ && $60.6$ & $60.1$ && $66.2$ & $61.3$  & $79.0$ && $77.3$ & $78.4$ \\
VQAScore (CLIP-FlanT5-XXL)~\cite{lin2024evaluating} & $\textbf{92.0}$ && $\textbf{91.6}$ & $\textbf{95.4}$ && $\underline{94.5}$ & $\underline{97.6}$  & $\textbf{88.3}$ && $\underline{93.5}$ & $82.9$ \\
\midrule
\rowcolor{gray!15}
\multicolumn{12}{c}{\textit{Encoder-based Methods}} \\
\midrule
CLIP-S/ViT-B~\cite{hessel2021clipscore} & $77.1$ && $78.0$ & $87.3$ && $82.6$ & $93.6$ & $69.1$ && $68.0$ & $61.2$ \\
CLIP-S/ViT-L~\cite{hessel2021clipscore} & $79.7$ && $83.0$ & $91.4$ && $84.9$ & $96.2$ & $70.8$ && $70.0$ & $61.6$ \\
LongCLIP-S/ViT-B~\cite{zhang2024long}   & $79.0$ && $76.0$ & $84.2$ && $85.7$ & $94.5$ & $74.8$ && $70.6$ & $67.4$ \\
LongCLIP-S/ViT-L~\cite{zhang2024long}   & $82.3$ && $86.4$ & $91.5$ && $86.9$ & $96.4$ & $76.8$ && $68.2$ & $70.2$ \\
PAC-S/ViT-B~\cite{sarto2023positive}    & $77.2$ && $75.3$ & $88.3$ && $82.0$ & $93.5$ & $72.6$ && $68.2$ & $60.8$ \\
PAC-S/ViT-L~\cite{sarto2023positive}    & $80.4$ && $83.0$ & \underline{91.8} && $86.0$ & $96.3$ & $76.3$ && $68.5$ & $60.8$ \\
PAC-S++/ViT-B~\cite{sarto2025positive}  & $80.5$ && $80.9$ & $90.6$ && $85.0$ & $95.3$ & $74.5$ && $70.6$ & $66.5$ \\
PAC-S++/ViT-L~\cite{sarto2025positive}  & $82.2$ && $86.9$ & $94.9$ && $87.4$ & \textbf{97.8} & $77.3$ && $67.7$ & $63.7$ \\
SPECS & $63.9$ && $82.1$ & $38.4$ && $70.7$ & $79.7$  & $57.3$ && $60.1$ & $58.8$ \\
\midrule
BEiTScore-B/32 w/o Stage 1 & $73.6$ && $62.3$ & $74.4$ && $71.7$ & $84.9$ & $65.4$ && $76.1$ & $69.4$ \\
BEiTScore-L/14 w/o Stage 1 & $88.3$ && $72.8$ & $84.0$ && $84.1$ & $92.7$ & $76.7$ && $89.6$ & $83.3$ \\
BEiTScore-B/32 & $81.9$ && $76.2$ & $78.3$ && $81.9$ & $90.1$ & $77.9$ && $89.0$ & $80.0$ \\
BEiTScore-L/14 & $88.7$ && $82.8$ & $86.9$ && $89.0$ & $93.8$ & $85.9$ && \textbf{94.0} & \underline{88.6} \\
\bottomrule
\end{tabular}}
\caption{Comparison against state-of-the-art metrics across the SugarCrepe~\cite{hsieh2023sugarcrepe} benchmark. Methods are categorized as encoder-based or LLM-based. Performance is measured using accuracy, and Avg-M stands for macro averaged accuracy. The highest score is highlighted using \textbf{bold}, and the second best is highlighted using an \underline{underline}. The results for models marked with $\dagger$ were taken from the original papers.}
\label{tab:sugarcrepe}
\vspace{-5mm}
\end{table}
To evaluate the ability of the models to distinguish hallucinations beyond nouns, we use the SugarCrepe~\cite{hsieh2023sugarcrepe} benchmark. This dataset presents models with ``hard negatives'' across three distinct tasks: the add task introduces new objects or attributes, to test sensitivity to over-description; the replace task substitutes correct words for new plausible but incorrect alternatives; finally, the swap task interchanges existing elements within a caption without disturbing the coherence of the sentence. According to the benchmark's authors, the swap task is considered the most challenging because it requires a deep understanding of structural and relational composition, rather than simple keyword matching.

The results in Table~\ref{tab:sugarcrepe} demonstrate that both the BEiTScore-B/32 and BEiTScore-L/14 variants surpasses existing encoder-based models on average, with BEiTScore-L/14 achieving performance comparable to LLM-based methods, despite being significantly smaller and more computationally efficient. Notably, our approach significantly outperforms encoder-based competitors on swap hard negatives, emphasizing the deeper understanding of structural and relational composition. Furthermore, ablation results confirm that Stage 1 consistently improves performance across all tasks on both size variants.

\subsection{Visio-Linguistic Compositional Perception}

Thus far, our evaluation has focused on the ability to identify hallucinations and perturbations involving individual nouns and attributes. In this subsection, we shift our focus to more complex textual perturbations that challenge the capacity for holistic compositional reasoning, rather than relying on shallow bag-of-words heuristics. To assess these advanced capabilities, we evaluate our models on two different benchmarks, namely VALSE~\cite{parcalabescu2022valse} and Winoground~\cite{thrush2022winoground}.

\subsubsection{Results on the VALSE Benchmark:}
\begin{table}[t]
\centering
\small
\resizebox{0.94\columnwidth}{!}{%
\begin{tabular}{l c c c c c c c c c c c}
\toprule
 & & \multicolumn{2}{c}{\textbf{Action}} && \multicolumn{3}{c}{\textbf{Counting}} & \multicolumn{4}{c}{\textbf{Counting}} \\
 \cmidrule(lr){3-4} \cmidrule(lr){5-8} \cmidrule(lr){9-12}
 & Avg-M. & Actant-swap & REPL && ADV & Balanced & SNS & Existence & Foil-it & Plurals & SPREL \\
\midrule
\rowcolor{gray!15}
\multicolumn{12}{c}{\textit{LLM-based Methods}} \\
\midrule
FLEUR~\cite{lee2024fleur}    & \underline{$87.7$} & $87.7$ & $87.7$ && $\textbf{88.0}$ & $\textbf{85.8}$  & $85.9$ & $87.1$ & $91.2$ & \textbf{86.4} & \underline{$90.5$} \\
EXPERT~\cite{kim2025expert}  & $85.2$ & $90.1$ & $\textbf{90.1}$ && $82.2$ & $80.4$  & $82.4$ & $84.0$ & $91.7$ & $\underline{85.2}$ & $\textbf{91.8}$ \\
GPT4o (API)    & $71.4$ & $\textbf{94.4}$ & $73.5$ && $63.1$ & $55.7$  & $55.9$ & $51.5$ & $84.9$ & $75.7$ & $86.0$ \\
VQAScore (CLIP-FlanT5-XXL)~\cite{lin2024evaluating} & $85.6$ & $82.3$ & $85.3$ && $80.0$ & $83.1$  & $\underline{87.9}$ & $95.0$ & $96.8$ & $78.1$ & $82.2$ \\
\midrule
\rowcolor{gray!15}
\multicolumn{12}{c}{\textit{Encoder-based Methods}} \\
\midrule
CLIP-S/ViT-B~\cite{hessel2021clipscore} & $64.2$ & $59.8$ & $82.3$ && $56.4$ & $64.2$ & $66.4$ & $74.6$ & $89.0$ & $63.1$ & $60.6$ \\
CLIP-S/ViT-L~\cite{hessel2021clipscore} & $66.0$ & $62.6$ & $85.2$ && $58.2$ & $62.0$ & $66.0$ & $73.9$ & $92.5$ & $66.3$ & $63.4$ \\
LongCLIP-S/ViT-B~\cite{zhang2024long}   & $69.1$ & $73.8$ & $82.4$ && $61.7$ & $68.8$ & $69.1$ & $69.1$ & $91.9$ & $66.6$ & $63.2$ \\
LongCLIP-S/ViT-L~\cite{zhang2024long}   & $68.6$ & $72,5$ & $\underline{89.0}$ && $60.2$ & $54.8$ & $58.1$ & $42.8$ & $96.9$ & $72.4$ & $68.6$ \\
PAC-S/ViT-B~\cite{sarto2023positive}    & $69.1$ & $65.1$ & $63.6$ && $69.5$ & $67.7$ & $67.3$ & $71.7$ & $69.9$ & $69.1$ & $75.7$ \\
PAC-S/ViT-L~\cite{sarto2023positive}    & $69.5$ & $64.3$ & $63.4$ && $69.5$ & $68.6$ & $66.9$ & $72.3$ & $77.8$ & $69.8$ & $78.1$ \\
PAC-S++/ViT-B~\cite{sarto2025positive}  & $69.7$ & $66.3$ & $66.2$ && $69.5$ & $68.9$ & $70.1$ & $74.9$ & $73.8$ & $69.7$ & $77.0$ \\
PAC-S++/ViT-L~\cite{sarto2025positive}  & $73.8$ & $72.0$ & $67.0$ && $69.2$ & $74.7$ & $73.8$ & $74.3$ & $78.3$ & $73.7$ & $78.1$ \\
SPECS~\cite{chen2025specs}    & $66.6$ & $80.0$ & $69.8$ && $79.7$ & $52.5$  & $47.7$ & $55.6$ & $89.2$ & $63.7$ & $60.9$ \\
12-in-1$\dagger$~\cite{parcalabescu2022valse}    & $75.7$ & $58.9$ & $65.9$ && $77.3$ & $76.7$ & $80.2$ & $\underline{95.6}$ & $86.9$ & $72.4$ & $67.7$ \\
\midrule
BEiTScore-B/32 without Stage 1 & $78.5$ & $85.8$ & $72.1$ && $72.7$ & $73.4$ & $80.1$ & $88.7$ & $94.7$ & $67.3$ & $71.8$ \\
BEiTScore-L/14 without Stage 1 & $87.1$ & $90.3$ & $83.0$ && $83.5$ & $83.0$ & $90.1$ & $97.0$ & $98.0$ & $75.2$ & $83.6$ \\
BEiTScore-B/32 & $82.3$ & $90.4$ & $81.6$ && $66.0$ & $73.9$ & $79.4$ & $94.1$ & $\underline{97.5}$ & $75.1$ & $83.0$ \\
BEiTScore-L/14 & $\textbf{89.2}$ & $\underline{93.9}$ & $86.7$ && \underline{$83.2$} & \underline{$85.7$} & $\textbf{88.8}$ & $\textbf{96.2}$ & $\textbf{99.2}$ & $82.5$ & $86.7$ \\
\bottomrule
\end{tabular}}
\caption{Comparison against state-of-the-art metrics across the VALSE~\cite{parcalabescu2022valse} benchmark. Methods are categorized as encoder-based or LLM-based. Performance is measured using accuracy. Avg-M stands for macro average accuracy, SNS for small numbers, ADV for adversarial, REPL for action replacement, SPREL for spatial relations. The highest score is highlighted using \textbf{bold}, and the second best is highlighted using an \underline{underline}. The results for models marked with $\dagger$ were taken from the original paper.}
\label{tab:valse}
\vspace{-5mm}
\end{table}
We evaluated both enconder-based and LLM-based models across all the tasks within the VALSE benchmark, with the exception of coreference, as it is the only task that lacks caption-like fluency. The results in Table~\ref{tab:valse} demonstrate that our BEiTScore-L/14 variant surpasses existing models on average, including the LLM-based EXPERT model that was fine-tuned on human judgments, despite our approach being significantly smaller and more computationally efficient. Furthermore, the BEiTScore-B/32 variant exceeds the encoder-based counterparts and achieves performance comparable to state-of-the-art LLM-based methods.

Compared to previous encoder-based models, including the best-performing model reported in the original benchmark paper, our method achieves higher performance on object-centric tasks such as existence and Foil-It. While the original authors suggested these tasks were nearly resolved by prior methods, due to their ability to correctly identify named objects, the experimental results show that our approach establishes a higher performance ceiling. 

Beyond these foundational tasks, our method also consistently achieves high performance on more challenging adversarial settings. Our models excel at distinguishing references to single versus multiple objects and counting them (plurality and counting), correctly classifying named spatial relations between objects (relations), and distinguishing actions while identifying their participants (actions). These tasks were previously very challenging for encoder-based methods, as highlighted in the original benchmark and confirmed by our results.

\subsubsection{Results on the Winoground Benchmark:}
\begin{figure}[t]
\centering
\begin{minipage}[t]{0.52\textwidth}
    \vspace{0pt}
    \centering
    
\centering
\small
\resizebox{\columnwidth}{!}{%
\begin{tabular}{l c c c}
\toprule
  & \textbf{Group} & \textbf{Text} & \textbf{Image} \\
\midrule
Human$\dagger$~\cite{thrush2022winoground}    & $85.5$ & $89.5$ & $88.5$ \\
Random Chance$\dagger$~\cite{thrush2022winoground}  & $16.7$ & $25.0$ & $25.0$ \\
\midrule
\rowcolor{gray!15}
\multicolumn{4}{c}{\textit{LLM-based Methods}} \\
\midrule
FLEUR~\cite{lee2024fleur}    & $28.8$ & $\underline{40.0}$ & $45.3$ \\
EXPERT~\cite{kim2025expert}  & $28.3$ & $40.0$ & $\underline{46.0}$ \\
VQAScore (CLIP-FlanT5-XXL) ~\cite{lin2024evaluating}    & $\textbf{44.8}$ & $\textbf{60.8}$ & $\textbf{55.3}$ \\
\midrule
\rowcolor{gray!15}
\multicolumn{4}{c}{\textit{Encoder-based Methods}} \\
\midrule
CLIP-S/ViT-B~\cite{hessel2021clipscore} & $8.5$ & $35.5$  & $11.3$ \\
CLIP-S/ViT-L~\cite{hessel2021clipscore} & $9.5$ & $30.8$ & $11.5$ \\
LongCLIP-S/ViT-B~\cite{zhang2024long}   & $7.3$ & $31.0$  & $9.0$ \\
LongCLIP-S/ViT-L~\cite{zhang2024long}   & $9.3$ & $29.8$  & $11.5$ \\
PAC-S/ViT-B~\cite{sarto2023positive}    & $6.8$ & $27.0$  & $10.0$ \\
PAC-S/ViT-L~\cite{sarto2023positive}    & $7.8$ & $26.0$  & $11.0$ \\
PAC-S++/ViT-B~\cite{sarto2025positive}  & $6.5$ & $31.8$  & $8.8$ \\
PAC-S++/ViT-L~\cite{sarto2025positive}  & $9.8$ & $31.5$  & $15.0$ \\
FLAVA$_{ITM}\dagger$~\cite{thrush2022winoground}  & $14.3$ & $32.3$  & $20.5$ \\
VinVL$\dagger$~\cite{thrush2022winoground}  & $14.5$ & $37.8$  & $17.$ \\
SPECS~\cite{chen2025specs}    & $3.0$ & $9.3$ & $7.8$ \\
\midrule
BEiTScore-B/32 without Stage1 & $14.3$ & $33.3$ & $22.5$ \\
BEiTScore-L/14 without Stage1 & $27.0$ & $48.0$  & $35.3$ \\
BEiTScore-B/32 & $18.3$ & $39.5$ & $28.5$ \\
BEiTScore-L/14 & $\underline{34.5}$ & $\underline{55.5}$  & $43.3$ \\
\bottomrule
\end{tabular}}
    \captionof{table}{Comparison against state-of-the-art reference-free metrics across the Winoground~\cite{thrush2022winoground}. Methods are categorized as encoder-based or LLM-based. The highest score is highlighted using \textbf{bold}, and the second best is highlighted using an \underline{underline}. The results for models marked with $\dagger$ were taken from the original paper.}
    \label{tab:winoground}
\end{minipage}
\hfill
\begin{minipage}[t]{0.45\textwidth}
    \vspace{0pt}
    \includegraphics[width=\linewidth]{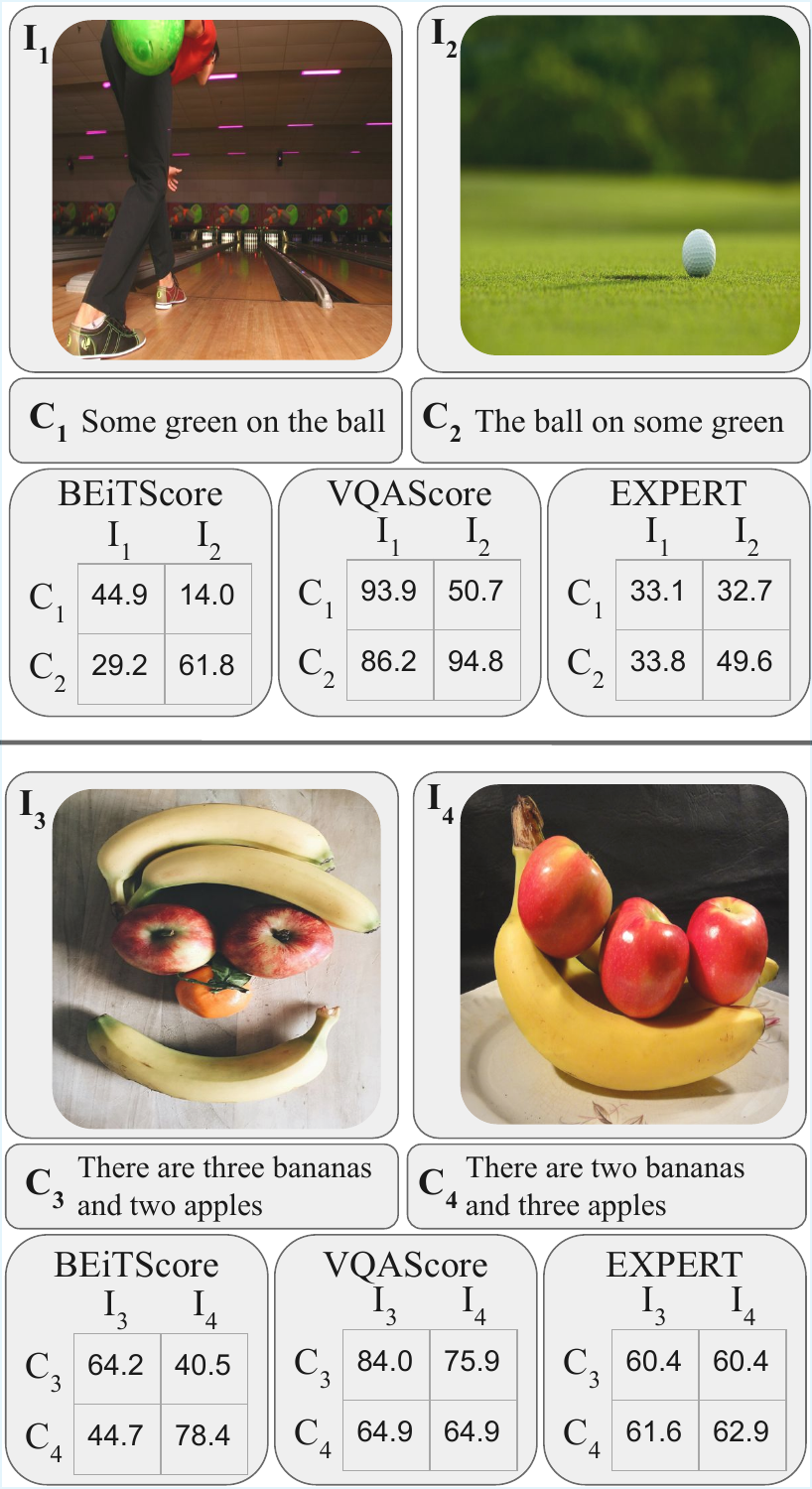}
    \caption{Two qualitative examples from Winoground benchmark. }
\end{minipage}

\end{figure}
Winoground features carefully hand-curated instances from expert annotators, designed for evaluating the ability of vision and language models to conduct visio-linguistic compositional reasoning. Given two images and two captions, the goal is to match them correctly. Crucially, both captions contain a completely identical set of words, only in a different order. Performance on Winoground is computed according to three different metrics that evaluate different aspects of a models’ visio-linguistic reasoning abilities. Text score measures whether a model can select the correct caption, given an image. Image score measures whether a model can select the correct image, given a caption. Group score combines the previous two, where every combination for a given example must be correctly scored by the model in order for the example to be considered correct. 

The results in Table~\ref{tab:winoground} highlight the superior performance of our models in visio-linguistic compositional reasoning. Notably, our BEiTScore-L/14 variant outperforms encoder-based methods across all metrics ny a significant margin. In addition, in surpasses LLM-based methods like EXPERT and FLEUR, as well as the best-performing encoder-based models cited in the original benchmark, such as FLAVA$_{ITM}$ and VinVL. It is important to highlight that both the BEiTScore-B/32 and BEiTScore-L/14 variants consistently exceed random-chance performance, a threshold that many previous encoder-based models struggle to meet. These results show that our approach effectively bridges the gap between efficient encoder-based architectures and the complex reasoning capabilities typically reserved for much larger and computationally heavy LLM-based frameworks.

\subsection{Detailed Visio-Linguistic Assessments Beyond 77-tokens}

\begin{figure}[t]
\centering
\begin{minipage}[t]{0.60\textwidth}
    \vspace{0pt}
    \vspace{1mm}
    \centering
    \centering
\small
\resizebox{\columnwidth}{!}{%
\begin{tabular}{l c c c c c c}
\toprule
  & \multicolumn{2}{c}{\textbf{Micro Avg.}} & \multicolumn{2}{c}{\textbf{VLCR}} & \multicolumn{2}{c}{\textbf{Scene Text}} \\
\cmidrule(lr){2-3} \cmidrule(lr){4-5} \cmidrule(lr){6-7} 
 & ACC & AUC & ACC & AUC & ACC & AUC \\
\midrule
\rowcolor{gray!15}
\multicolumn{7}{c}{\textit{LLM-based Methods}} \\
\midrule
FLEUR~\cite{lee2024fleur}    & $83.5$ & $68.1$ & $82.4$ & $66.5$ & $87.9$ & $79.1$ \\
EXPERT~\cite{kim2025expert}  & $\textbf{93.4}$ & $\underline{70.8}$ & $\textbf{94.3}$ & $\underline{70.3}$ & $\underline{89.8}$ & $\underline{82.3}$ \\
GPT4o (API) & $77.1$ & $--$ & $80.8$ & $--$ & $62.3$ & $--$ \\
VQAScore (CLIP-FlanT5-XXL)~\cite{lin2024evaluating} & $87.5$ & $\textbf{77.2}$ & $86.1$ & $\textbf{72.1}$ & $\textbf{93.3}$ & $\textbf{91.7}$ \\
\midrule
\rowcolor{gray!15}
\multicolumn{7}{c}{\textit{Encoder-based Methods}} \\
\midrule
CLIP-S/ViT-B~\cite{hessel2021clipscore} & $16.7$ & $54.9$ & $0.0$ & $50.0$ & $81.7$ & $74.7$ \\
CLIP-S/ViT-L~\cite{hessel2021clipscore} & $17.9$ & $55.7$ & $0.0$ & $50.0$ & $87.8$ & $78.2$ \\
LongCLIP-S/ViT-B~\cite{zhang2024long}   & $70.9$ & $59.7$ & $66.7$ & $54.5$ & $87.4$ & $75.0$ \\
LongCLIP-S/ViT-L~\cite{zhang2024long}   & $72.6$ & $57.6$ & $69.2$ & $54.0$ & $85.9$ & $72.3$ \\
PAC-S/ViT-B~\cite{sarto2023positive}    & $19.1$ & $53.8$ & $0.0$ & $50.0$ & $79.2$ & $68.7$ \\
PAC-S/ViT-L~\cite{sarto2023positive}    & $20.3$ & $55.2$ & $0.0$ & $50.0$ & $85.8$ & $76.0$ \\
PAC-S++/ViT-B~\cite{sarto2025positive}  & $19.0$ & $53.5$ & $0.0$ & $50.0$ & $79.2$ & $67.6$ \\
PAC-S++/ViT-L~\cite{sarto2025positive}  & $19.9$ & $54.8$ & $0.0$ & $50.0$ & $86.8$ & $74.1$ \\
SPECS~\cite{chen2025specs}              & $70.5$   & $54.6$   & $71.8$  & $56.8$   & $65.7$   & $55.8$ \\
\midrule
BEiTScore-B/32 without Stage 1 & $66.3$ & $55.1$ & $64.6$ & $53.8$ & $72.7$ & $60.5$ \\
BEiTScore-L/14 without Stage 1 & $76.1$ & $60.5$ & $73.7$ & $59.3$ & $85.3$ & $70.1$ \\
BEiTScore-B/32 & $85.4$ & $67.3$ & $85.5$ & $66.6$ & $84.9$ & $73.7$ \\
BEiTScore-L/14 & $\underline{90.5}$ & $68.0$ & $\underline{90.7}$ & $68.5$ & $\underline{89.8}$ & $\underline{82.3}$ \\
\bottomrule
\end{tabular}}

    \captionof{table}{Results across LongCapVLCP. Methods are categorized as encoder-based or LLM-based. Performance is measured using Accuracy. The highest score is highlighted using \textbf{bold}, and the second best is highlighted using an \underline{underline}.}
    \label{tab:longcapvlcr}
\end{minipage}
\hfill
\begin{minipage}[t]{0.37\textwidth}
    \vspace{0pt}
    \includegraphics[width=\linewidth]{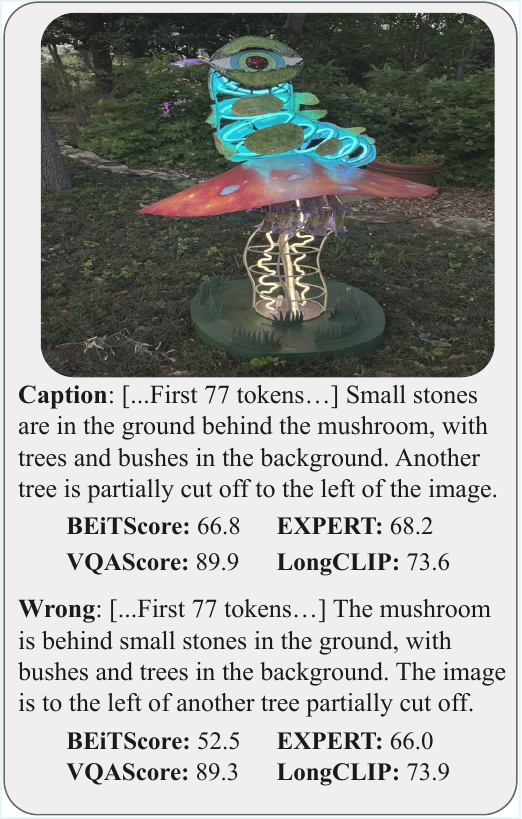}
    \caption{Qualitative example from LongCapVLCP, with an Image and original caption from DOCCI.}
\end{minipage}
\end{figure}
We reported results across existing benchmarks that provide a robust framework for assessing holistic compositional perception through textual perturbations. However, current benchmarks focus on captions under 77-tokens or position critical discriminative errors early in the text, failing to challenge a model’s ability to reason on long narratives. To bridge this gap, we introduce LongCapVLCP to test visio-linguistic compositional perception beyond the 77-token threshold. Furthermore, we broaden the evaluation scope by examining whether existing metrics can effectively deal with scene text. Specifically, we test the model's ability to distinguish between a ground-truth caption containing the correct textual reference and a corrupted version containing an incorrect textual reference, requiring the metric to accurately map the visual text within the image to its corresponding mention in the caption.

The results presented in Table~\ref{tab:longcapvlcr} show that standard encoder-based metrics, such as CLIP-S or PAC-S, are ineffective for this task, performing no better than random chance. This stems from the fact that our benchmark places textual corruptions exclusively after the 77-token threshold (with the exception of the scene text instances), i.e. a region that falls outside the fixed context window of these models. In contrast, both variants of the proposed metric significantly surpass the LongCLIP and FLEUR models on average. Notably, our BEiTScore-L/14 model achieves performance on par with the LLM-based EXPERT model, demonstrating its ability to maintain discriminative capabilities throughout long captions without the computational overhead of a large language model. Regarding the analysis of scene text, BEiTScore-L/14 outperforms encoder-based methods, with a performance comparable to LLM-based metrics. The consistent performance gap between our models with and without Stage 1 training highlights the clear benefits of the multi-stage training curriculum, for both the scene text and long captioning contexts.

\section{Conclusions}

In this paper, we introduced BEiTScore as a new reference-free image captioning evaluation metric that bridges the gap between efficient encoder-based architectures and computationally intensive LLM-based frameworks. Our extensive evaluation across eleven benchmarks demonstrates that BEiTScore not only achieves state-of-the-art correlation with human judgments, surpassing larger LLM-based methods like EXPERT, FLEUR and VQAScore, but also excels in identifying complex hallucinations and caption corruptions where previous encoder-based methods frequently failed.

Another contribution of this work is the introduction of the LongCapVLCP benchmark, which evaluates vision-linguistic compositional perception for long captions with respect to a broad spectrum of visual–textual relations, including scene text recognition an aspect largely underrepresented in prior work. 

Our findings reveal that traditional CLIP-based models fail to process information beyond the 77-token threshold. In contrast, BEiTScore maintains strong compositional reasoning in long-form narratives, and strong scene text understanding abilities, matching or exceeding the performance of specialized LLM-based metrics while remaining significantly more efficient. These results support our hypothesis: the recent dominance of LLM-based metrics is not merely a product of scale, but a reflection of the poor exposure of smaller models to more complex and long-form textual data during training. Ultimately, BEiTScore provides a cost-effective solution for automated evaluation, offering strong correlation with human judgments and intricate compositional perception.

%
%
\bibliographystyle{splncs04}
\bibliography{main}
\newpage

\appendix
\section*{Supplementary Materials}
We prepared a set of supplementary materials that provide additional details supporting the methodology and results presented in the main paper. We first present a statistical characterization of the pairwise datasets used for training and validating our models, as well as for the newly introduced LongCapVLCP benchmark, including caption length distributions and dataset composition (see Section \ref{app:datasets}). Next, we describe the prompts employed for generating incorrect captions through LLM-based augmentation in Section \ref{app:prompts}. We then provide a more detailed breakdown of the results on the LongCapVLCP benchmark across different error categories (Section \ref{app:longcap}). Finally, in Section \ref{app:genai} we report extended evaluation results on the GenAI-Bench~\cite{li2024evaluating} benchmark, offering further insights into the performance of the proposed metrics compared with existing state-of-the-art approaches to assess the quality of AI-generated imagery.

\section{Statistical Characterization of the Pairwise Datasets}
\label{app:datasets}
\begin{table}[b!]
\centering
\resizebox{\columnwidth}{!}{%
\begin{tabular}{lcccc}
\toprule
Type of Augmentation & ~FOIL-IT~ & ~Localized Narratives~ & ~PixelProse~ & ~TextCaps~ \\
\midrule
\textbf{Training Split (Total)} & 348934 & 386467 & 226764 & 98083 \\
\quad Foil & 187841 & 90537 & 87314 & 0 \\
\quad Action & 34004 & 59886 & 25502 & 0 \\
\quad Nouns & 34004 & 59048 & 25502 & 0 \\
\quad Relations & 34004 & 59344 & 25502 & 0 \\
\quad Actant & 34004 & 58996 & 25502 & 0 \\
\quad Counting & 25077 & 58656 & 37442 & 0 \\
\quad Scene Text & 0 & 0 & 0 & 98083 \\
\quad Number of Unique Images & 65201 & 240607 & 173666 & 21948 \\
\midrule
\textbf{Validation Split (Total)} & 35920 & 31581 & 31260 & 7361 \\
\quad Foil & 9886 & 4749 & 9702 & 0 \\
\quad Action & 6000 & 5194 & 4500 & 0 \\
\quad Nouns & 6000 & 5210 & 4500 & 0 \\
\quad Relations & 6000 & 5266 & 4500 & 0 \\
\quad Actant & 6000 & 5290 & 4500 & 0 \\
\quad Counting & 2034 & 5872 & 3558 & 0 \\
\quad Scene Text & 0 & 0 & 0 & 7361 \\
\quad Number of Unique Images & 19218 & 20625 & 22191 & 2748 \\
\bottomrule
\end{tabular}}
\caption{Dataset composition for the pairwise training and validation splits. We report the number of instances and number of unique images in the training and validation splits across datasets and semantic categories. Each instance corresponds to a triplet, featuring an image paired with a correct caption and an incorrect caption.}
\label{tab:train_val_splits}
\vspace{-5mm}
\end{table}

\begin{table}[t]
\centering
\resizebox{0.69\columnwidth}{!}{%
\begin{tabular}{lccc}
\toprule
Type of Augmentation & ~~IWW~~ & ~~DOCCI~~ & ~~TextCaps~~ \\
\midrule
\textbf{LongCapVLCP (Total)} & 1620 & 713 & 597 \\
\quad Action & 307 & 131 & 0 \\
\quad Nouns & 318 & 146 & 0 \\
\quad Relations & 322 & 139 & 0 \\
\quad Actant & 302 & 137 & 0 \\
\quad Counting & 371 & 160 & 0 \\
\quad Scene Text & 0 & 0 & 597 \\
\quad Number of Unique Images & 381 & 99 & 555 \\
\bottomrule
\end{tabular}}
\caption{Dataset composition for the LongCapVLCP benchmark. We report on the number of instances, where each instance corresponds to an image paired with a correct caption and an incorrect caption.}
\label{tab:longcapvlcp}
\vspace{-5mm}
\end{table}

\begin{figure}[b!]
    \centering
    \includegraphics[width=\textwidth]{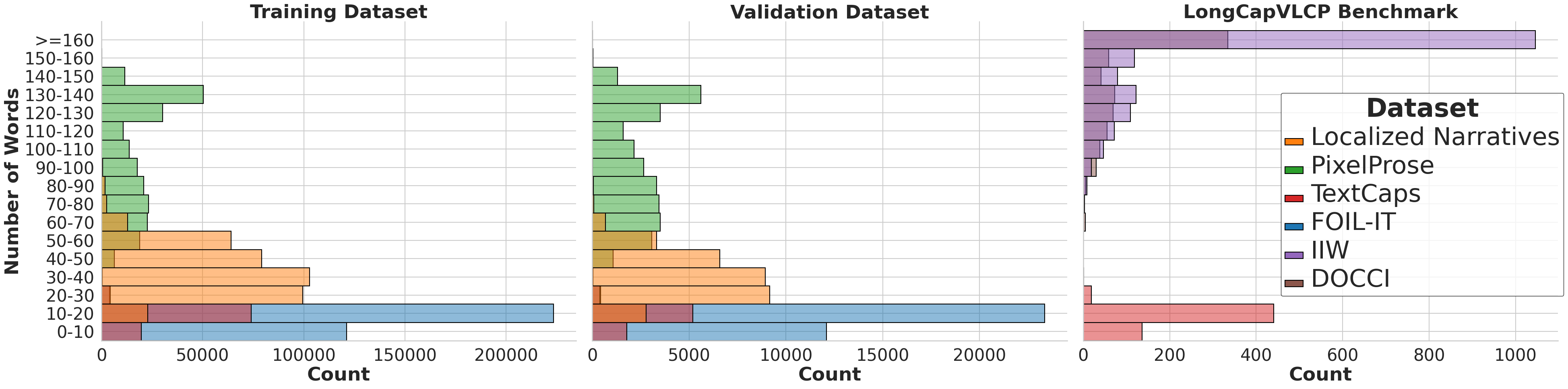}  
    \caption{Caption length distributions (i.e., the number of words per caption) for the three datasets. Each histogram shows the number of instances containing captions with a number of words that falls within each bin interval values. The maximum caption lengths are as follows: $176$ words for the training dataset, $270$ words for the validation dataset, and $525$ words for the LongCapVLCP benchmark.}
    \label{fig:caption_histograms}
    \vspace{-5mm}
\end{figure}

This section provides a statistical characterization of the pairwise datasets used for training, validation, and benchmarking. Although the training strategy uses two stages corresponding to pairwise classification and regression data based on human judgments, the later stage corresponds to the use of well-established datasets that have already been described in detail in prior work (and hence these other datasets are not presented here in detail).

Tables~\ref{tab:train_val_splits} and~\ref{tab:longcapvlcp} summarize the number of instances across the original source datasets, and across semantic categories corresponding to the type of intervention used to create the negative captions. Table~\ref{tab:train_val_splits} specifically covers the pairwise datasets used for training and validation, while Table~\ref{tab:longcapvlcp} describes the LongCapVLCP benchmark dataset.

Each of the instances in the counts that are reported in both tables corresponds to a triplet, featuring an image paired with its ground-truth caption and an incorrect negative caption.  Besides presenting the number of instances/triples, Table~\ref{tab:train_val_splits} also features rows with the total number of images per split. In the instances/triples, the incorrect caption is either generated through the LLM-augmented strategy or obtained directly from the original dataset (as in the case of the Foil-It dataset used in the training and validation splits). In addition, two other types of negative captions per instance are used during training but are not reflected in the tables, since they do not correspond to fixed negatives. First, synthetically corrupted captions are produced through word or sentence shuffling, which are generated on demand during training. Second, retrieval-based negative captions are obtained through negative mining and organized into three difficulty levels (easy, medium, and hard), each containing up to 30 candidate captions per level per instance.

When forming contrastive pairs during training, a negative caption is randomly sampled from the aforementioned pool of negative candidates. This sampling strategy maintains a balanced ratio between correct and incorrect instances, while exposing the model to a diverse range of semantic errors.

For the LongCapVLCP benchmark, only negatives generated through the LLM-augmented strategy are included. Therefore, the numbers of images, positive captions, and negative captions in the benchmark, are all fixed and correspond directly to the number of instances reported in the table.

Figure \ref{fig:caption_histograms} shows the distribution of caption lengths, measured as the number of words per caption, across our datasets. The first histogram illustrates the distribution for the training split, the second for the validation split, and the third for the LongCapVLCP benchmark. These plots highlight the differences in caption length across datasets and splits, providing insight into the variability and complexity of the textual descriptions used for model training and evaluation. Notice that LongCapVLCP features several captions with a much larger length than that of the training captions.

\section{Prompts}
\label{app:prompts}

This section presents the prompts used to generate incorrect captions through LLM-based augmentation. The boxes in red correspond to prompts for the Qwen3 model, used for generating hard negative examples for training data. The boxes in blue correspond to prompts for GPT4o, used to generate negative captions for the LongCapVLCP benchmark.

\begin{redprompt}[title= Qwen3 Prompt to Generate Incorrect Captions With Noun/Attribute Errors]
Rewrite the following image caption by changing some of the nouns or adjectives in the text to incorrect alternatives. The original caption is factually correct, but the rewritten caption should be factually incorrect by replacing nouns or adjectives with different ones, that subtly change the meaning of the text. Modify only certain elements, such as nouns for objects, adjectives that quality the objects, colors, materials, object attributes, background, scenarios, or general settings, while keeping most of the original caption and of its structure unchanged. Apply a sparse and varied mix of changes within each caption. The result should sound natural and coherent, while being factually incorrect. Preserve the sentence structure and overall fluency. If it is not possible to rewrite the caption by changing nouns or adjectives while maintaining fluency and structure, output exactly the original caption. Do not include any explanations or additional text—only output besides the rewritten incorrect caption or the original caption. 

\textbf{<caption>:\{correct\_caption\}}
\end{redprompt}

\begin{redprompt}[title= Qwen3 Prompt to Generate Incorrect Captions With Actant Swapping Errors]
Write a wrong and incoherent version of the following original image caption, by changing the roles of the agents by the objects in the sentence. Actant swapping means swapping the subject and object roles between each other in the text. For instance, the sentence 'The boy sits in the table' should be replaced by 'The table sits on the boy'. In turn,  'The women is holding a book' should be replaced by 'The book is holding the woman'. As another example, the sentence 'The apple is on the table' should be replaced by 'The table is on the apple'. The rewritten caption should be factually incorrect, by reversing agent and object roles when this is possible. If it is not possible to rewrite the caption by swapping these roles while maintaining fluency and structure, output exactly the original caption. Do not include any explanations or additional text—only output besides the rewritten incorrect caption or the original caption. 

\textbf{<caption>:\{correct\_caption\}}
\end{redprompt}

\begin{redprompt}[title= Qwen3 Prompt to Generate Incorrect Captions With Action Verbs Errors]
Rewrite the following image caption by changing some of the action verbs in the sentence to incorrect ones. The original caption is factually correct, but the rewritten caption should be factually incorrect by replacing several of the action verbs with different ones, this way changing the meaning of the caption. Preserve the overall caption structure and overall fluency. If it is not possible to rewrite the caption by changing the action verbs while maintaining fluency and structure, output exactly the original caption. Do not include any explanations or additional text—only output besides the rewritten incorrect caption or the original caption. 

\textbf{<caption>:\{correct\_caption\}}
\end{redprompt}

\begin{redprompt}[title= Qwen3 Prompt to Generate Incorrect Captions With Spatial Relation Errors]
Rewrite the following image caption, while preserving the overall caption structure and fluency. The resulting caption should contain errors regarding spatial relationships between the elements being described. The resulting caption should be grammatically fluent, but containing several spatial relationship errors, for instance obtained by replacing spatial prepositions in the original caption. If it is not possible to rewrite the caption by changing spatial relationships while maintaining fluency and structure, output exactly the original caption. Do not include any explanations or additional text—only output besides the rewritten incorrect caption or the original caption. 

\textbf{<caption>:\{correct\_caption\}}
\end{redprompt}

\begin{blueprompt}[title= GPT4o Prompt to Generate Incorrect Captions With Spatial Relation Errors]
Based on the input image and the following image caption, write a wrong caption with respect to spatial relationships only. Preserve the overall structure and fluency of the original caption. The resulting caption should be grammatically fluent, but containing spatial relationship errors, for instance obtained by replacing spatial prepositions in the original caption. If it is not possible to rewrite the caption by changing spatial relationships while maintaining fluency and structure, output exactly the word False. Do not include any explanations or additional text—only output besides the rewritten incorrect caption or the word False. 

\textbf{<correct\_caption>:\{correct\_caption\}}
\end{blueprompt}

\begin{blueprompt}[title= GPT4o Prompt to Generate Incorrect Captions With Actant Swapping Errors]
Based on the input image and the following image caption, write a wrong caption with respect to swapping the subject and object roles between each other in the text. Write a wrong and incoherent version of the original caption by changing the roles of the agents by the object in the text. Actant swapping means swapping the subject and object roles between each other in the text. For instance, the sentence 'The boy sits in the table' should be replaced by 'The table sits on the boy'. In turn, 'The women is holding a book' should be replaced by 'The book is holding the woman'. As another example, tOhe sentence ´The apple is on the table' should be replaced by 'The table is on the apple'. The rewritten caption should be factually incorrect by reversing the agent and object roles when this is possible. If it is not possible to rewrite the caption by swapping these roles, while maintaining fluency and structure, output exactly the word False. Do not include any explanations or additional text—only output besides the rewritten incorrect caption or the word False. 

\textbf{<correct\_caption>:\{correct\_caption\}}
\end{blueprompt}

\begin{blueprompt}[title= GPT4o Prompt to Generate Incorrect Captions With Action Verb Errors]
Based on the input image and the following image caption, write a wrong caption with respect to action verbs only. Rewrite the caption by changing some of the action verbs in the sentence to incorrect ones. The original caption is factually correct, but the rewritten caption should be factually incorrect by replacing some of the action verbs with different ones, changing the meaning of the text. Preserve the caption structure and overall fluency. If it is not possible to rewrite the caption by changing the action verbs while maintaining fluency and structure, output exactly the word False. Do not include any explanations or additional text—only output besides the rewritten incorrect caption or the word False. 

\textbf{<correct\_caption>:\{correct\_caption\}}
\end{blueprompt}

\begin{blueprompt}[title= GPT4o Prompt to Generate Incorrect Captions With Counting Errors]
Based on the input image and the following image caption, write a wrong caption with respect to numeric quantifiers only. Preserve the caption structure and overall fluency. The resulting caption should contain errors regarding numeric quantifiers, such as incorrect counts or amounts. The resulting caption should be grammatically fluent, although containing numerical quantifier errors. If it is not possible to rewrite the caption by changing numeric quantifiers while maintaining fluency and structure, output exactly the word False. Do not include any explanations or additional text—only output besids the rewritten incorrect caption or the word False.

\textbf{<correct\_caption>:\{correct\_caption\}}
\end{blueprompt}

\begin{blueprompt}[title= GPT4o Prompt to Generate Incorrect Captions With Noun/Attribute Errors]
Based on the input image and the following image caption, write a wrong caption with respect to changing nouns and adjectives in the text to incorrect ones. Rewrite the caption by changing some of the nouns or adjectives in the text to incorrect ones. The original caption is factually correct, but the rewritten caption should be factually incorrect by replacing nouns or adjectives with different ones, that subtly change the meaning of the sentence. Only modify elements such as nouns, adjectives, colors, materials, attributes, background, scenarios, or settings, while keeping most of the caption unchanged. Apply a sparse and varied mix of changes to the caption. The result should sound natural and coherent while being factually incorrect. Preserve the overall structure and fluency. If it is not possible to rewrite the caption by changing nouns and adjectives, while maintaining fluency and structure, output exactly the word False. Do not include any explanations or additional text—only output besides the rewritten incorrect caption or the word False. 

\textbf{<correct\_caption>:\{correct\_caption\}}
\end{blueprompt}

\section{LongCapVLCP Results in More Detail}
\label{app:longcap}
\begin{table}[t!]
\centering
\small
\resizebox{\columnwidth}{!}{%
\begin{tabular}{l c c c c c c c c c c c c c c }
\toprule
  & \multicolumn{2}{c}{\textbf{Micro Avg.}} & \multicolumn{2}{c}{\textbf{Actant-Swap}} & \multicolumn{2}{c}{\textbf{Actions}} & \multicolumn{2}{c}{\textbf{Counting}} & \multicolumn{2}{c}{\textbf{Nouns}} & \multicolumn{2}{c}{\textbf{SPREL}} & \multicolumn{2}{c}{\textbf{Scene Text}} \\
\cmidrule(lr){2-3} \cmidrule(lr){4-5} \cmidrule(lr){6-7} \cmidrule(lr){8-9} \cmidrule(lr){10-11} \cmidrule(lr){12-13} \cmidrule(lr){14-15}
 & ACC & AUC & ACC & AUC & ACC & AUC & ACC & AUC & ACC & AUC & ACC & AUC & ACC & AUC\\
\midrule
\rowcolor{gray!15}
\multicolumn{15}{c}{\textit{LLM-based Methods}} \\
\midrule
FLEUR~\cite{lee2024fleur}    & $83.5$ & $68.1$ & $77.0$ & $\underline{64.6}$ & $83.3$ & $66.0$ & $82.5$ & $61.7$ & $89.4$ & $75.2$ & $79.4$ & $65.8$ & $87.8$ & $79.1$\\
EXPERT~\cite{kim2025expert}  & $\textbf{93.5}$ & $\underline{70.8}$ & $\underline{91.3}$ & $\textbf{68.9}$ & $\underline{94.8}$ & $70.6$ & $\textbf{95.1}$ & $\underline{65.9}$ & $\textbf{97.2}$ & $\underline{78.5}$ & $\textbf{93.1}$ & $\underline{68.0}$ & $\underline{90.0}$ & $\underline{82.3}$\\
GPT4o (API) & $77.1$ & $--$ & $\textbf{95.0}$ & $--$ & $87.2$ & $--$ & $64.6$ & $--$ & $85.3$ & $--$ & $75.5$ & $--$ & $62.3$ & $--$\\
VQAScore (CLIP-FlanT5-XXL)~\cite{lin2024evaluating} & $87.5$ & $\textbf{77.2}$ & $76.5$ & $63.0$ & $87.7$ & $\underline{71.5}$ & $85.7$ & $\textbf{68.2}$ & $94.4$ & $\textbf{84.0}$ & $85.7$ & $\textbf{73.7}$ & $\textbf{93.3}$ & $\textbf{91.7}$\\
\midrule
\rowcolor{gray!15}
\multicolumn{15}{c}{\textit{Encoder-based Methods}} \\
\midrule
CLIP-S/ViT-B~\cite{hessel2021clipscore} & $16.7$ & $54.9$ & $0.0$ & $50.0$ & $0.0$ & $50.0$ & $0.0$ & $50.0$ & $0.0$ & $50.0$ & $0.0$ & $50.0$ & $81.7$ & $74.7$\\
CLIP-S/ViT-L~\cite{hessel2021clipscore} & $17.9$ & $55.7$ & $0.0$ & $50.0$ & $0.0$ & $50.0$ & $0.0$ & $50.0$ & $0.0$ & $50.0$ & $0.0$ & $50.0$ & $87.8$ & $78.2$\\
LongCLIP-S/ViT-B~\cite{zhang2024long}   & $70.9$ & $59.7$ & $56.7$ & $52.7$ & $67.6$ & $53.9$ & $59.5$ & $51.6$ & $81.7$ & $60.4$ & $68.6$ & $54.1$ & $87.4$ & $75.0$\\
LongCLIP-S/ViT-L~\cite{zhang2024long}   & $72.6$ & $57.6$ & $58.8$ & $52.0$ & $65.1$ & $53.0$ & $65.7$ & $52.2$ & $86.4$ & $59.1$ & $69.6$ & $53.9$ & $85.9$ & $72.3$\\
PAC-S/ViT-B~\cite{sarto2023positive}    & $19.1$ & $53.8$ & $0.0$ & $50.0$ & $0.0$ & $50.0$ & $0.0$ & $50.0$ & $0.0$ & $50.0$ & $0.0$ & $50.0$ & $79.2$ & $68.7$\\
PAC-S/ViT-L~\cite{sarto2023positive}    & $20.3$ & $55.2$ & $0.0$ & $50.0$ & $0.0$ & $50.0$ & $0.0$ & $50.0$ & $0.0$ & $50.0$ & $0.0$ & $50.0$ & $85.8$ & $76.0$\\
PAC-S++/ViT-B~\cite{sarto2025positive}  & $19.0$ & $53.5$ & $0.0$ & $50.0$ & $0.0$ & $50.0$ & $0.0$ & $50.0$ & $0.0$ & $50.0$ & $0.0$ & $50.0$ & $79.2$ & $67.6$\\
PAC-S++/ViT-L~\cite{sarto2025positive}  & $19.9$ & $54.8$ & $0.0$ & $50.0$ & $0.0$ & $50.0$ & $0.0$ & $50.0$ & $0.0$ & $50.0$ & $0.0$ & $50.0$ & $86.8$ & $74.1$\\
SPECS~\cite{chen2025specs} & $70.5$ & $54.6$ & $71.8$ & $59.2$ & $71.5$ & $57.1$ & $64.0$ & $51.7$ & $83.2$ & $59.9$ & $69.4$ & $57.1$ & $65.7$ & $55.8$\\
\midrule
BEiTScore-B/32 without Stage 1 & $66.3$ & $55.1$ & $52.4$ & $50.1$ & $71.7$ & $53.9$ & $64.8$ & $53.7$ & $71.8$ & $57.1$ & $61.6$ & $53.9$ & $72.7$ & $60.5$\\
BEiTScore-L/14 without Stage 1 & $76.1$ & $60.5$ & $59.2$ & $53.8$ & $66.4$ & $57.0$ & $78.9$ & $58.2$ & $86.4$ & $67.8$ & $75.7$ & $59.1$ & $85.3$ & $70.1$\\
BEiTScore-B/32 & $85.3$ & $67.3$ & $72.9$ & $60.2$ & $92.5$ & $70.3$ & $86.8$ & $64.2$ & $92.5$ & $74.6$ & $82.4$ & $64.2$ & $84.9$ & $73.7$\\
BEiTScore-L/14 & $\underline{90.5}$ & $68.0$ & $81.8$ & $63.8$ & $\textbf{95.4}$ & $\textbf{71.8}$ & $\underline{91.9}$ & $65.4$ & $\underline{97.0}$ & $75.9$ & $\underline{86.8}$ & $66.1$ & $89.8$ & $\underline{82.3}$\\
\bottomrule
\end{tabular}}
\caption{Results across LongCapVLCP categories. Methods are categorized as encoder-based or LLM-based. Performance is measured using accuracy and Area Under the ROC curve (AUC). SPREL stands for spatial relationships. The highest score is highlighted using \textbf{bold}, and the second best is highlighted using an \underline{underline}.}
\label{tab:longcapvlcp_detail}
\vspace{-5mm}
\end{table}
This section provides a detailed analysis of each captioning error type included in the LongCapVLCP benchmark: actant swapping, action verbs, numeric quantifiers, nouns and attributes, spatial relationships, and scene text. Consistent with the earlier overall analysis, standard encoder-based metrics such as CLIP-S and PAC-S are ineffective over this benchmark, performing no better than random chance. In contrast, both variants of the proposed metric substantially outperform LongCLIP and FLEUR, on average. Notably, the BEiTScore-L/14 model achieves performance comparable to the LLM-based EXPERT metric, demonstrating an ability to preserve discriminative power across long captions, without incurring the computational overhead associated with large language models.

In the specific case of instances involving scene text, BEiTScore-L/14 also surpasses other encoder-based approaches and reaches performance levels similar to those of LLM-based metrics. Furthermore, several trends previously observed in other datasets remain evident. For instance, GPT-4o performs surprisingly well at detecting actant swap errors, while most models consistently identify noun and attribute errors with relatively high accuracy. Overall, the proposed model achieves highly competitive performance compared with LLM-based metrics across the different error categories.

\section{Results on the GenAI-Bench Dataset}
\label{app:genai}
In this section, we evaluate the performance of existing image captioning metrics using GenAI-Bench~\cite{li2024evaluating} to assess how well both our proposed method and other state-of-the-art automated evaluation metrics align with human perception in the context of AI-generated imagery. According to the authors of the benchmark, although image captioning evaluation models have demonstrated strong performance on photorealistic imagery, evaluating adherence to complex and compositional prompts requires metrics capable of capturing intricate logic, attributes, and spatial relationships. To address this challenge, GenAI-Bench was constructed by prompting several state-of-the-art image generation models to produce images from professional-grade textual prompts. Each resulting image–caption pair was then evaluated by human experts. These expert ratings serve as the ground-truth baseline for measuring the correlation between automated metric scores and human judgments. The results comparing our proposed models with state-of-the-art evaluation metrics are presented in Table~\ref{tab:genai}.

The results demonstrate that both BEiTScore-B/32 and BEiTScore-L/14 outperform existing encoder-based methods, on average, with BEiTScore-L/14 achieving performance comparable to LLM-based approaches, despite being significantly smaller and more computationally efficient. Notably, these results indicate that our models can effectively process and evaluate image generation outputs, exhibiting performance comparable to LLM-based strategies. This additional experiment highlights the feasibility of using our approach also as a reliable and efficient metric for assessing the quality of AI-generated imagery.

\begin{table}[t]
\centering
\small
\begin{tabular}{l c c c}
\toprule
\textbf{Model} &\textbf{Kendall $\tau_c$} & \textbf{Kendall $\tau_b$} & \textbf{Spearman $\rho$} \\
\midrule
\rowcolor{gray!4}
\multicolumn{4}{c}{\textit{LLM-based Methods}} \\
\midrule
FLEUR~\cite{lee2024fleur}   & 34.0 & 33.5 & 44.5 \\
EXPERT~\cite{kim2025expert} & \underline{35.6} & \underline{35.1} & 46.3 \\
VQAScore (CLIP-FlanT5-XXL)~\cite{lin2024evaluating} & \textbf{38.4} & \textbf{37.9} & \textbf{49.9} \\
\midrule
\rowcolor{gray!4}
\multicolumn{4}{c}{\textit{Encoder-based Methods}} \\
\midrule
CLIP-S/ViT-B~\cite{hessel2021clipscore} & 10.0 & 9.9 & 13.3 \\
CLIP-S/ViT-L~\cite{hessel2021clipscore} & 11.8 & 11.6 & 15.6 \\
LongCLIP-S/ViT-B~\cite{zhang2024long}   & 13.4 & 13.2 & 17.7 \\
LongCLIP-S/ViT-L~\cite{zhang2024long}   & 15.0 & 14.8 & 19.8\\
PAC-S/ViT-B~\cite{sarto2023positive}    & 12.9 & 12.7 & 17.1  \\
PAC-S/ViT-L~\cite{sarto2023positive}    & 12.2 & 12.1 & 16.3  \\
PAC-S++/ViT-B~\cite{sarto2025positive}  & 14.6 & 14.4 & 19.3 \\
PAC-S++/ViT-L~\cite{sarto2025positive}  & 16.3 & 16.1 & 21.6 \\
SPECS~\cite{chen2025specs}              & 1.5 & 1.4 & 1.9 \\
\midrule
BEiTScore-B/32 w/o Stage 1 & 28.0 & 27.6 & 36.7 \\
BEiTScore-L/14 w/o Stage 1 & 33.1 & 32.7 & 43.2 \\
BEiTScore-B/32             & 33.1 & 32.7 & 43.2  \\
BEiTScore-L/14             & \underline{35.6} & \underline{35.1} & \underline{46.5} \\
\bottomrule
\end{tabular}
\caption{Comparison against state-of-the-art reference-free metrics on the GenAI-Bench dataset. Methods are categorized as encoder-based or LLM-based. Performance is measured using Kendall $\tau_c$, Kendall’s $\tau_b$, and Spearman $\rho$ correlation with human judgments. The highest score is highlighted using \textbf{bold}, and the second best is highlighted using an \underline{underline}.}
\label{tab:genai}
\vspace{-5mm}
\end{table}

\end{document}